\documentclass[sigconf]{acmart}

\AtBeginDocument{%
  \providecommand\BibTeX{{%
    \normalfont B\kern-0.5em{\scshape i\kern-0.25em b}\kern-0.8em\TeX}}}

\setcopyright{acmcopyright}
\copyrightyear{2018}
\acmYear{2018}
\acmDOI{10.1145/1122445.1122456}

\acmConference[Woodstock '18]{Woodstock '18: ACM Symposium on Neural
  Gaze Detection}{June 03--05, 2018}{Woodstock, NY}
\acmBooktitle{Woodstock '18: ACM Symposium on Neural Gaze Detection,
  June 03--05, 2018, Woodstock, NY}
\acmPrice{15.00}
\acmISBN{978-1-4503-XXXX-X/18/06}

\begin{document}

\title{Remember What You have drawn: Semantic Image Manipulation with Memory}

\author{Xiangxi Shi}
\authornote{indicates equal contribution.}
\affiliation{%
\institution{Electrical Engineering and Computer Science}
  \institution{Oregon state university}
  \country{}
  }
\email{shixia@oregonstate.edu}

\author{Zhonghua Wu}
\authornotemark[1]
\affiliation{%
\institution{School of Computer Science and Engineering}
  \institution{Nanyang Technological University}
  \country{}
  }
\email{zhonghua001@e.ntu.edu.sg}
  
\author{Guosheng Lin}
\affiliation{%
  \institution{School of Computer Science and Engineering}
  \institution{Nanyang Technological University}
  \country{}
  }
\email{gslin@ntu.edu.sg}

\author{Jianfei Cai}
\affiliation{%
    \institution{Department of Data Science \& AI}
  \institution{Monash University}
  \country{}
  }
\email{Jianfei.Cai@monash.edu}

\author{Shafiq Joty}
\affiliation{%
    \institution{School of Computer Science and Engineering}
  \institution{Nanyang Technological University}
  \country{}
  }
\email{srjoty@ntu.edu.sg}

\begin{abstract}
Image manipulation with natural language, which aims to manipulate images with the guidance of language descriptions, has been a challenging problem in the fields of computer vision and natural language processing (NLP). Currently, a number of efforts have been made for this task, but their performances are still distant away from generating realistic and text-conformed manipulated images. Therefore, in this paper, we propose a memory-based Image Manipulation Network (MIM-Net), where a set of memories learned from images is introduced to synthesize the texture information with the guidance of the textual description. We propose a two-stage network with an additional reconstruction stage to learn the latent memories efficiently. To avoid the unnecessary background changes, we propose a Target Localization Unit (TLU) to focus on the manipulation of the region mentioned by the text. Moreover, to learn a robust memory, we further propose a novel randomized memory training loss. Experiments on the four popular datasets show the better performance of our method compared to the existing ones.
\end{abstract}




\maketitle

\begin{figure}[h]
\centering
    \includegraphics[width=0.9\linewidth]{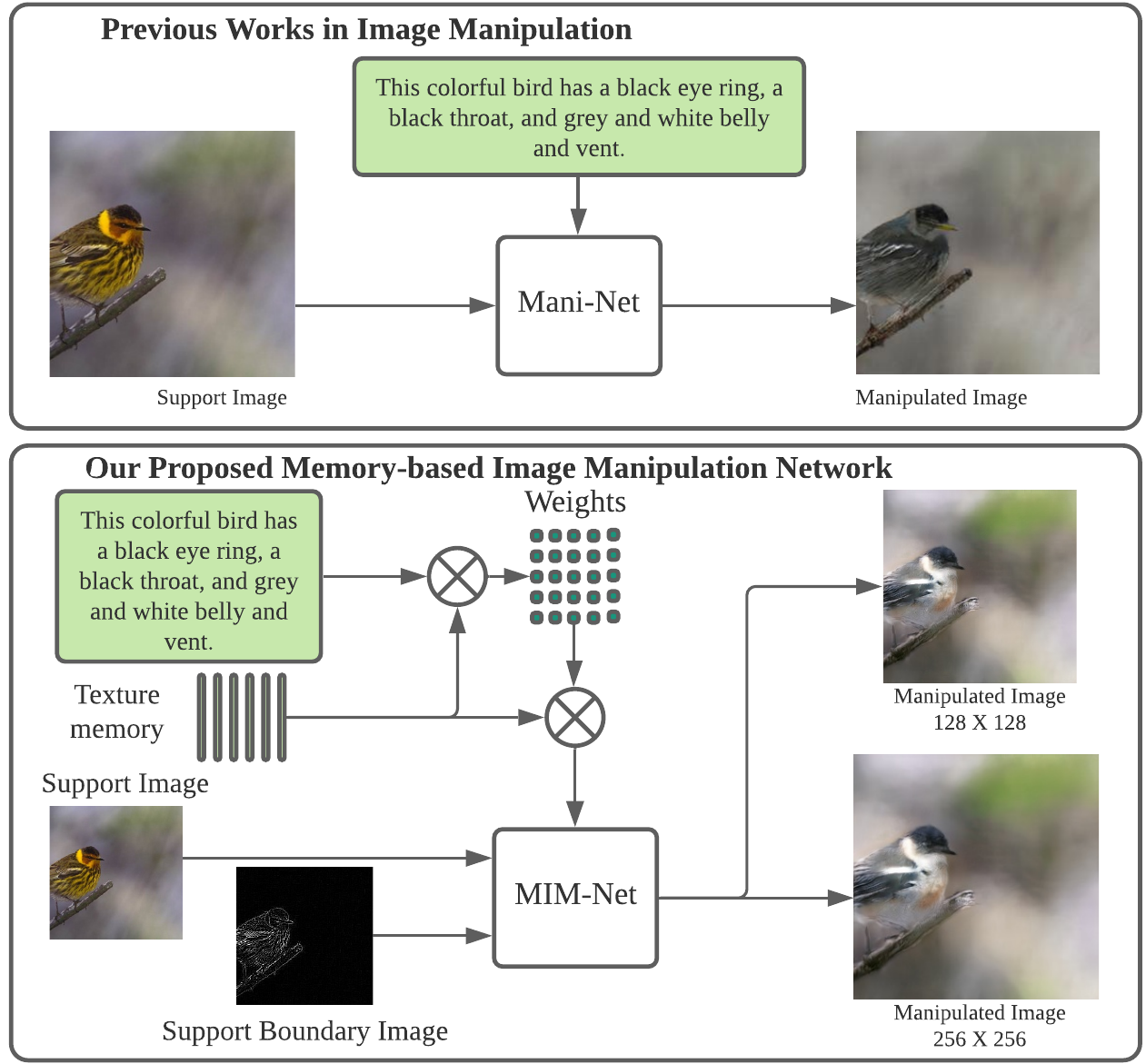}
    \caption{Illustration of the image manipulation task.}
    \label{first_diagram}
\end{figure}

\section{Introduction}

Images are one of the most important content types on the Internet. There is a growing requirement for images to be edited according to personal preferences.  
However, the image processing software (e.g., Photoshop) usually requires professional techniques and complicated steps to do the required manipulation. It is highly desirable to do automatic image manipulation  according to a natural language description so as to help people who have little knowledge about professional software, as exemplified in the top part of Fig. \ref{first_diagram}.

\begin{figure}[t]
\centering
    \includegraphics[width=0.7\linewidth]{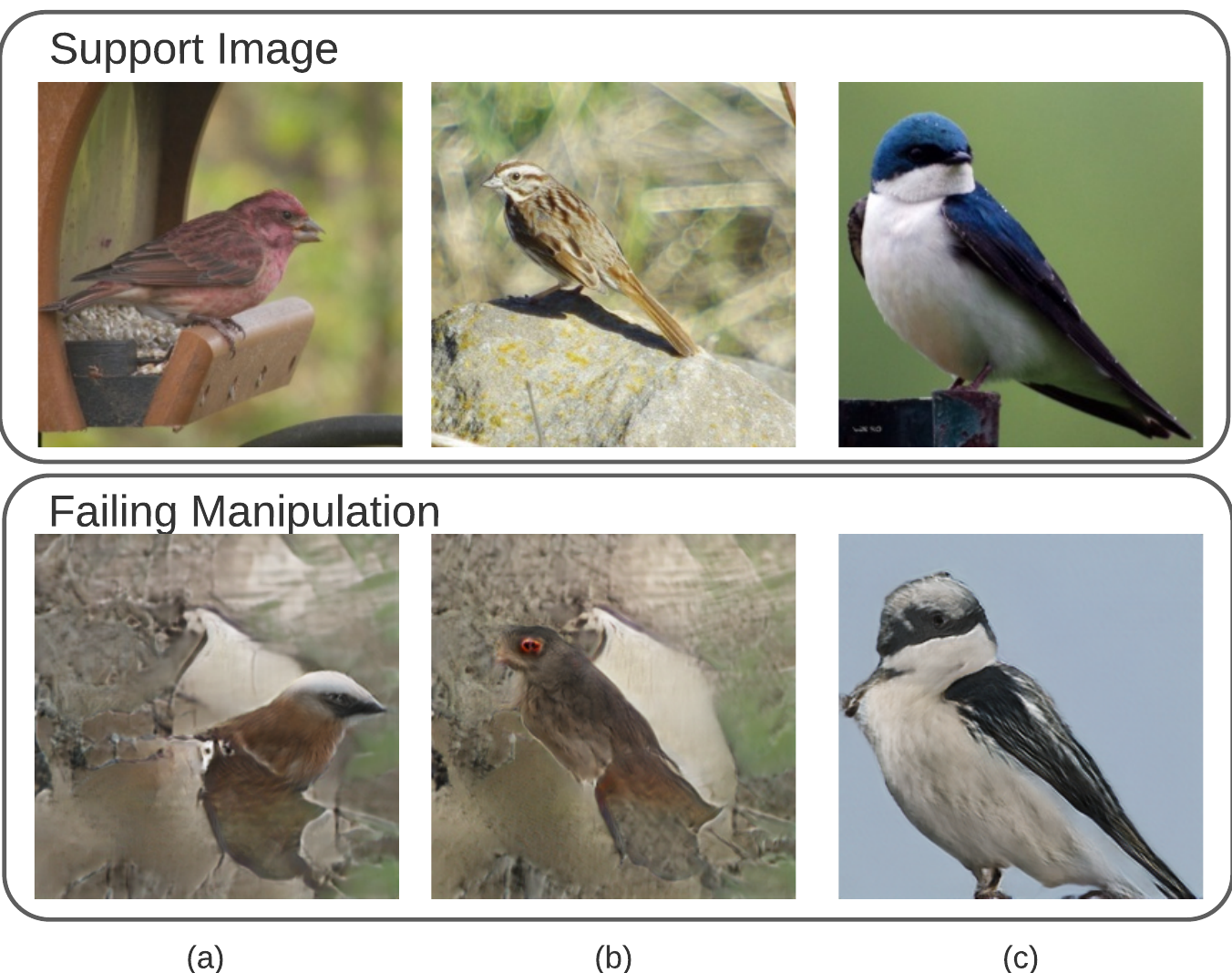}
    \caption{Failing samples generated from ManiGAN \cite{li2020manigan}}
    \label{failingcase}
\end{figure}

Earlier language-based image modification network \cite{nam2018tagan,dong2017semantic} usually takes the image and language as input to the network at the same time. The characteristic information of the image and language are respectively processed through the image and language encoders. The feature information is then fused through a convolutional or transformation layer, and decoded into the modified picture. To get a better mapping between image and text features, more recent methods \cite{xu2018attngan,li2020manigan} pre-train encoders using the provided language and image pairs before training the manipulation network. These methods can help the model to get an initial mapping between images and texts. However, the pretrained encoders also bring some problems. As shown in Fig. \ref{failingcase}, due to overfitting in the encoder pretraining process, they tend to record irrelevant information like background.

In examples (a) and (b) in Fig.~\ref{failingcase}, although the support images are different, the model still generates quite similar backgrounds. 
Such effect is likely due to the pretrained language feature which contains some irrelevant information because of the overfitting on the dataset. In (c), the model modifies the background from green to blue which is 
the most frequent background in the dataset.

Besides the cross-modality gap, structural distortion is another problem observed in the previous works \cite{nam2018tagan,li2020manigan}. Because of the unique feature used to represent the attribute of the manipulated object, texture and structure information are blended together, which makes it easy to interfere with each other during the manipulation.

To overcome the problems mentioned above, in this work we propose a ``Memory based Image Manipulation Network'' (MIM-Net) with pretrained word embeddings to avoid structural distortion and bridge the gap between the two modalities at the same time. As shown in the lower part of Fig. \ref{first_diagram}, in our approach, we disentangle visual features into two parts: texture and structural information. We introduce a set of memories to remember the texture information, while boundary features are used to represent the structure information. To support a better cross-modal mapping, the corresponding image features retrieved from the memories are used instead of word embedding as the editing features. We propose a set of loss functions to keep the induced memories meaningful and robust. 
In addition, we introduce a novel target region estimation unit between the edited image features and memories to encourage the model to keep an accurate background and also learn better memories.

We evaluate our method on CelebA, CUB, Oxford-102, and DeepFashion datasets. It shows that our proposed MIM-Net is able to both manipulate the images based on the given text and preserve the background. The contributions of this paper are summarized as follows:

\begin{itemize}

    \item We propose a memory-based image manipulation method MIM-Net to learn a more representative texture feature for manipulation.
    
    \item A novel target localization unit (TLU) is proposed to let our MIM-Net focus on the manipulation of the region mentioned by the text. The proposed unit is able to help the memory network learn more valuable features.
    \item We propose compound loss functions and training strategies to make our memory module memorize valuable texture information.
    \item We demonstrate the effectiveness of our proposed MIM-Net on four data sets including Celeba, CUB, Oxford-102 and DeepFashion.
    
\end{itemize}

\section{Related Work}

 Generative Adversarial Networks (GAN)~\cite{goodfellow2014generative} has been very  successful in various image generation tasks. In a GAN, a generated image is distinguished by a discriminator to identify if it is real or not and force the generator to synthesize a realist image through back-propagation. In this section, we review relevant GAN based methods for different tasks.  

\textbf{Conditional Generative Adversarial Networks.}
With the development of GAN in the image generation field, Mirza et la. \cite{mirza2014conditional} proposed conditional GAN (cGAN) to generate images conditioned on auxiliary information. Chen et al.~\cite{chen2016infogan} proposed to learn a meaningful latent representation based on the information theories. Inspired by cGAN, Pix2Pix\cite{isola2017image} was introduced by Isola et al. to the image translation task, which aims to transfer the style of a real image from the source domain to the target domain. Later, CycleGAN\cite{zhu2017unpaired} was proposed for unpaired image translation by a cycled image translation between source and target domains. StyleGAN~\cite{karras2019style} was introduced to map the latent features into parameters that can be adapted in the adaptive instance normalization (AdaIN) for the style transformation.

\textbf{Text to Image Synthesis.} Considering the importance of the natural language in human interactions, several works \cite{reed2016generative,reed2016learning,zhang2017stackgan} were proposed focusing on generating an image based on a language description. Reed et al.~\cite{reed2016generative} proposed a GAN-based model which combines noise prior and description embedding as the input. Later, in~\cite{reed2016learning}, Reed et al. further proposed a location-sensitive method by involving object bounding boxes as auxiliary information. To generate high-quality images with a larger resolution, Zhang et al.\cite{zhang2017stackgan} introduced a 2-step method that first generates an image from the description and uses it as an additional input to generate a larger image in the second-step GAN model. To avoid the lack of word-level information due to the global sentence embedding, Xu et al.~\cite{xu2018attngan} proposed to apply attention modules to calculate the similarity between image patches and words.

\begin{figure*}[t]
    \centering
    \includegraphics[height=1.6in,width=0.8\linewidth]{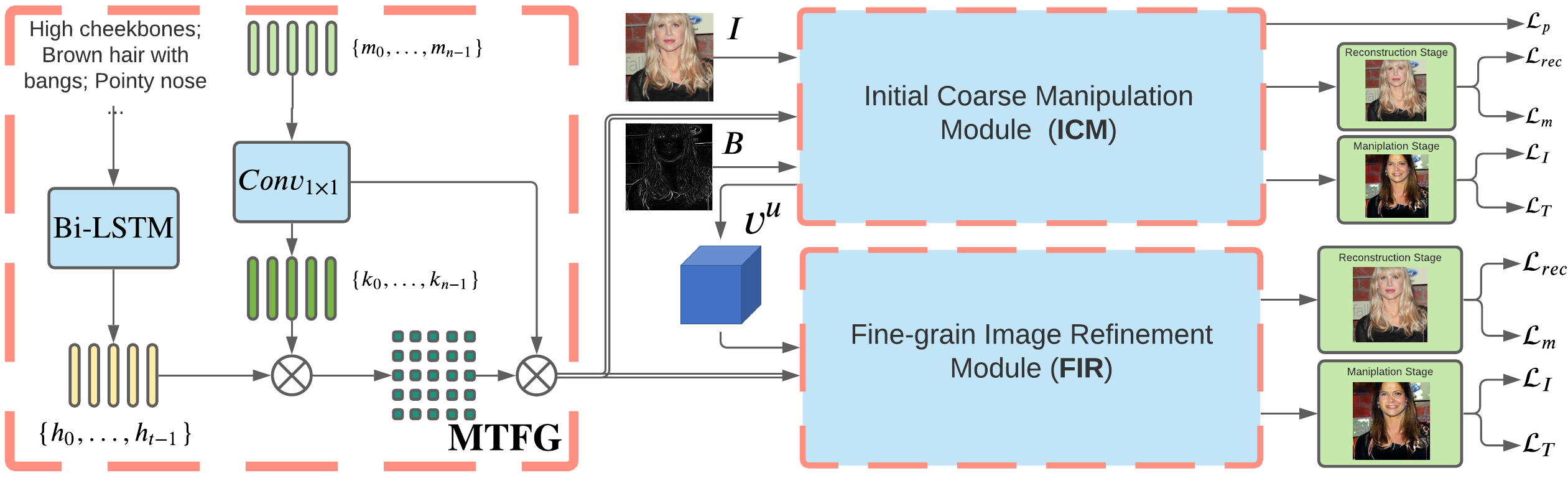}
    \caption{The overall architecture of our MIM-Net. The Memory-based Texture Feature Generation (MTFG) is the core module, which supports a two-stage image manipulation in coarse- and fine-grain levels through two generators: Initial Coarse Manipulation Module (ICM) and Fine-grain Image Refinement Module (FIR). $I$ is the input image and $B$ is the boundary image. $v^u$ is the upsampled manipulated feature map generated in ICM module. During training, the model is trained with $(\mathcal{L}_{rec}$, $\mathcal{L}_m)$ and $(\mathcal{L}_I$, $\mathcal{L}_T)$ respectively in the reconstruction and manipulation stages. ICM module also uses another loss $\mathcal{L}_p$ to learn robust memories. }
    \label{overall}
\end{figure*}

\textbf{Image Manipulation.} Different from the text-to-image synthesis task, image manipulation, introduced by Dong et al.~\cite{dong2017semantic}, aims to manipulate an image according to a given natural language description. In the paper, Dong et al. proposed a GAN-based model that combines image and language information at feature level in generator and discriminator to achieve image manipulation and evaluation. Nam et al.~\cite{nam2018text} proposed a word-level discriminator to overcome the mismatch between the descriptions and images caused by coarse training feedback from sentence-level discriminators. Li et al.~\cite{li2020manigan} proposed an affine combination module to merge the information of image and text for a better attribute generation. Liu et al.~\cite{liu2020ir} proposed an adversarial mechanism to figure out the semantic increment in each step for multi-step iterative manipulation.

\textbf{Image Generation with Memory.} To provide plentiful and meaningful information for the image generation, memory has been introduced to various image generation tasks \cite{tseng2020retrievegan,xu2020texture,qi2018semi,zhu2019dm,yoo2019coloring}. In \cite{qi2018semi,tseng2020retrievegan,xu2020texture}, memory banks, which are consisted of image patches related to generation requirements, are provided to the model to improve the reality and diversity of generated images. Yoo et al.~\cite{yoo2019coloring} proposed to learn a latent memory set in the GAN-base model for image recoloring tasks. Zhu et al.~\cite{zhu2019dm} updated the memory iteratively based on language feature and the memory from the last step to refine the synthetic images.

Different from these previous works, which manipulate the images with the given description directly, our method has an auxiliary reconstruction step to learn a memory set from the entire dataset for different attribute representations. To avoid the over-focusing on image features during reconstruction, we propose a novel target localization unit to preserve the manipulated maps in the reconstruction step and learn more robust attribute representations with it. Compared with the memory-based image generation methods~\cite{zhu2019dm,yoo2019coloring}, our method introduces a compound loss to keep the memory meaningful by generating an image with a randomly selected memory feature during training. With the learned memory, the model can reduce the gap between image and language by generating the manipulated representation from the retrieved memory feature instead of language embedding.

\section{Methods}

We propose a novel two-stage generation network, called MIM-Net. As shown in Fig. \ref{overall}, the network contains a memory based texture generation module (MTFG) that manipulates the image in two stages. In the first stage (ICM), we synthesize texture features of objects from the latent memory set with the guidance of the given textual description. Then a coarser level manipulated image with low resolution is generated from the given image and texture features. 
In the second stage (FIR), we further refine the image combining with the attentive texture features to a $256 \times 256$ resolution.

Since the ground-truth manipulated image is not available during training, both ICM and FIR generate the manipulated images with the help of two discriminators through adversarial training. Besides the adversarial training, we also introduce a reconstruction training process to learn a more representative and diverse memory set.

\subsection{Memory-based Texture Feature Generation} 

To model the texture information, we propose to learn a set of latent texture memories from the entire dataset, which allows the model to synthesize the texture features more closely to real image features. We denote the memory set as 
\begin{equation}
    \mathcal{M} = \{m_0, m_1, ..., m_{n-1}\},
\end{equation}
where $m_i \in \mathcal{R}^l$ is a $l$-dimensional memory vector and $n$ is the size of the memory set. A set of corresponding key features are calculated to enable the language descriptions to select the most relevant visual texture features from the memory set. In this way, our model can bridge the gap between image and textual description and avoid the noise from language resources.

 \begin{figure*}[t]
    \centering
    \includegraphics[width=0.8\linewidth]{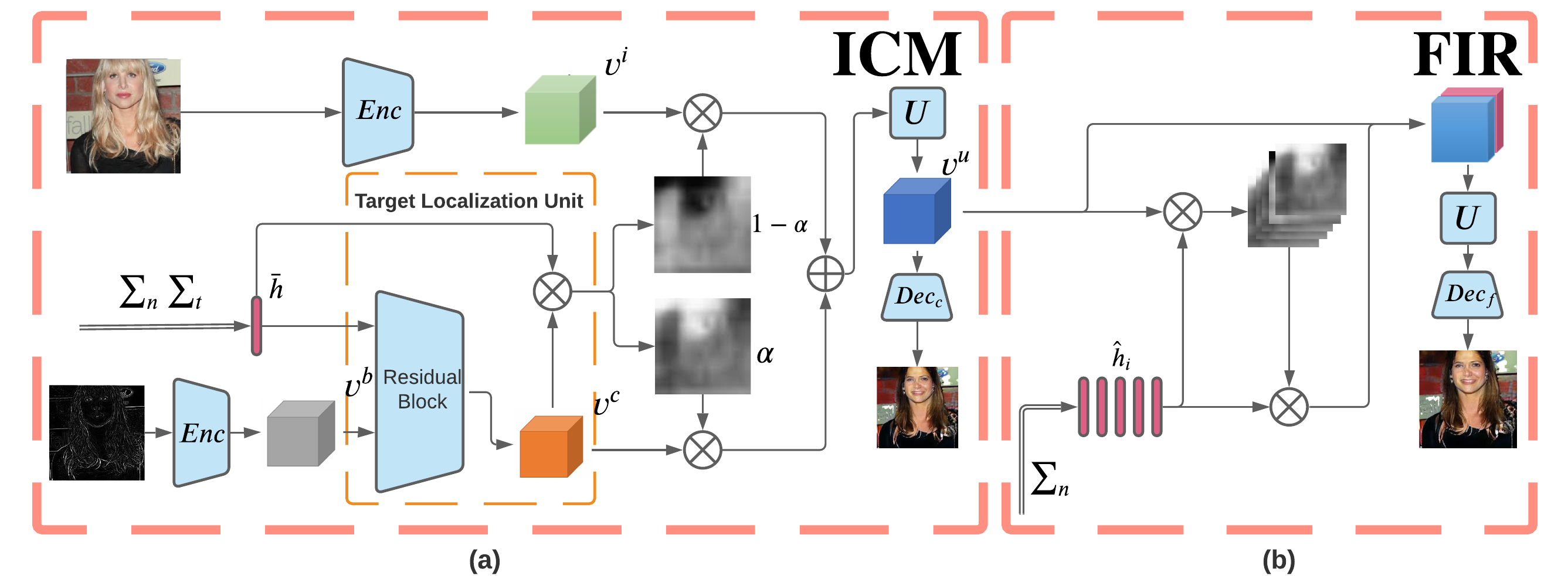}
    \caption{Illustration of our Initial Coarse Manipulation Module (ICM) and Fine-grain Image Refinement Module (FIR) modules. The structure in yellow box is the proposed Target Localization Unit (TLU). The ICM module takes the image feature $v^i$, the boundary feature $v^b$, and the global texture feature $\bar{h}$ as input to generate a coarse manipulated image. The FIR module takes word-level texture feature $\hat{h}_i$and upsampled feature map $v^u$ as the input to generated a refined image. $Enc$ is a pre-trained CNN network for feature extraction, and $Dec_c$ and $Dec_f$ are the decoders for the ICM and FIR respectively. $U$ is the upsampling function.}
    \label{generators}
\end{figure*} 

In the MTFG module, we first encode the language description into word-level text features with a Bi-directional Long Short-term Memory (Bi-LSTM), denoted as $\{h_0, h_1, \cdots, h_{t-1}\}$ with $t$ being the description length.
At the same time, a Linear projection matrix is applied to the memories to generate the keys for texture information in text domain. With the keys and word-level features, we can predict a weight map through a soft attention module and merge the memories into word-level texture features
\begin{align}
    a_{ij} &= \frac{exp(h_i\cdot Wm_j)}{\sum_{k = 0}^{n-1} exp(h_i\cdot Wm_k) }, \\
    \hat{h}_i &= \sum_{j = 0}^{n-1} a_{ij}m_j.
    \label{fuse}
\end{align}
where $W$ is the projection matrix, $Wm_j$ is the key feature, $a_{ij}$ is the attention weight between word-level feature and memory key, and $\hat{h}_i$ is the fused word-level texture feature.

\subsection{Initial Coarse Manipulation Module}
 
In the ICM stage, the network aims to generate a coarse manipulated image that generally matches the description (Fig. \ref{generators}(a)). For this, the word-level texture features are averaged to get a global feature to represent the information of the entire description.
In this stage, a visual feature map $v^b$, which is extracted from a boundary image by a pre-trained Convolutional Neural Network (CNN), is introduced to model the structure information of the input image. We concatenate the global texture feature with boundary features and pass it through a Residual Block $f_r(v) + v$, which consists of two convolutional layers with activation functions to synthesize the coarse manipulated features. 
 
To distinguish the background from the manipulated objects, we propose a novel target region estimation unit  called \textit{Target Localization Unit} (Sec. \ref{subsec:TLU}) to predict a weight map $\alpha$ (a matrix of the same size as $v^b$), which is used to fuse the manipulated features and original image features. Finally, we upsample the fused feature map to the resolution of $128\times128$ and feed it as input to a decoder (see the supplementary material for details) to generate the coarse image with the same resolution:
 \begin{align}
    \bar{h} &= \sum_{i = 0}^{t-1} \hat{h}_i, \\
    \label{h_i}
    v_{xy} &= W_r[v^b_{xy}, \bar{h}], \\
    \label{v^c}
    v^c &= f_r(v) + v, \\
    v^u &= U(\alpha \odot v^c + (1 - \alpha) \odot v^i), \\
    I_c &= Dec_c(v^u)
\end{align}
where $W_r$ is a projection matrix, $\bar{h}$ is the global texture feature, $v^c$ is the coarse manipulated feature map which contains the attributes of object after manipulation, $U$ is the upsampling function, and $\odot$ is the element-wise multiplication; $Dec_c$ is the decoder to decode the manipulated features to RGB images.

\subsection{Target Localization Unit} \label{subsec:TLU}

Different from the previous studies \cite{xu2018attngan,li2020manigan} which mainly apply attention methods between image and text features, we predict a weight map to capture the most texture-related region in the (to be) manipulated feature map. The weight map involves the global texture feature $\bar{h}$ and coarse manipulated feature map $v^c$ as the key and value respectively in the calculation. The formulation of the TLU unit is,
\begin{align}
    \alpha_{xy} &= \sigma(v^c_{xy}\cdot \bar{h}),
\end{align}
where $\sigma$ is the sigmoid function, and $v^c_{xy}$ is the $x$-th row and $y$-th column patch of the coarse manipulated feature. 

Since the coarse manipulated feature is generated from the texture and structure features through the residual block $f_r(v) + v$ (Eq. \ref{h_i} - \ref{v^c}), during training, the block is expected to preserve the texture information by having higher weight values ($\alpha_{xy}$) for the regions of $v^c$ that we intend to manipulate. On the other hand, it is encouraged to filter the background information from the manipulated feature map by having lower weight values for those regions. In this way, the network is expected to learn the intrinsic correlation between textures and structures by exploring the combination of the given texture and different structures in residual blocks.

With the weight map, the network can find the valid texture-structure pairs in the feature map and filter the rest that are treated as background in the adversarial training. Moreover, during the reconstruction stage, the manipulated regions would be preserved by the TLU. With a salient weight map, the model is able to learn more representative texture memories by reconstructing the objects with manipulated features instead of the original image features. 
 
\subsection{Fine-grain Image Refinement Module} 

In the ICM stage, the network generates a coarse image. However, because of the global texture feature, some fine grain texture attributes are covered by the general ones that lead to losing some local details. 

Inspired by \cite{zhu2019dm}, following the ICM stage, we further introduce a Fine-grain Image Refinement (FIR) stage to refine the image at a finer level (Fig. \ref{generators}(b)). In this stage, the model aims to refine the upsampled feature according to the word-level texture features. We first predict the similarity between the word-level texture features $\hat{h}_k$ and upsampled features $v^u_{xy}$. Then the network synthesizes a corresponding texture feature for each upsampled feature. After a channel-wise concatenation, we upsample the concatenated feature map to the resolution of $256\times256$ and decode it to the fine-grain image with the same resolution. More formally, 
\begin{align}
    h^f_{xy} &= \frac{1}{T}\sum_{k=0}^{t-1}\sigma(v^u_{xy}\cdot\hat{h}_k)\hat{h}_k \\
    I_f &= Dec_f(U([v^u, h^f]))
\end{align}
where $v^u_{xy}$ is the upsampled feature generated in the ICM stage, $h^f_{xy}$ is the texture feature for $x$-th row and $y$-th column patch of the upsampled image, and $I_f$ is the refined image. Similar to ICM, $Dec_f$ and $U$ are the decoder for fine-grain refinement and the upsampling function, respectively.

\subsection{Training}

\subsubsection{Discriminators}

Following \cite{nam2018tagan}, the discriminators consists of two branches evaluating the text-conformity and the reality of the manipulated images as
\begin{align}
    &D_I(I) = \sigma(f(I)) \\
    D_T(I, T) = &\prod^{t-1}_{i=0}[\sigma(f(I) \cdot W(h_i) + b(h_i))]^{\alpha_i} 
    \label{alfa}
\end{align}
where $D_I(I)$ is the reality discriminator and $D_T(I,T)$ is the text-conformity discriminator. $f(\cdot)$ is a CNN network to predict a feature for  the evaluated image, 
and $\sigma$ is the sigmoid function.
$h_i$ is the hidden state of a Bi-LSTM encoded from the $i$th word embedding. $W(\cdot)$ and $b(\cdot)$ are the projection functions and the bias to predict the evaluation score of the image feature. In Eq. \ref{alfa}, $\alpha_i \in [0, 1]$ is a similarity weight between $h_i$s and the average of them to predict the importance of each word-level information in the description. Then the $\alpha_i$ values are applied as exponents of the evaluation scores to highlight the important ones and reduce the influence of unimportant ones in the multiplication function.

\subsubsection{Reconstruction Training}

Because of the lack of ground truth after manipulation, the previous works train the model with adversarial technique, which supervises the model by neural-network based discriminators. Although such an unsupervised way can overcome the label-lacking problems in image manipulation task, the random data sampling is likely to bring in noise from the unpaired train data and lead to failure in learning accurate texture information. To learn more accurate and representative memories, apart from the adversarial generation stage, we introduce a reconstruction stage where the model manipulates images with its paired descriptions. Note that to keep memories accurate, they are only updated during this training stage.

\textbf{Reconstruction Loss.} We feed the images and their paired descriptions to the network and expect the model can reconstruct the input images. We introduce an $L_2$ loss instead of the adversarial losses in this stage, 
\begin{align}
    \mathcal{L}_{rec}^i = \mathbb{E}_{I,T}[||I - G^i(I, T)||_2],
\end{align}
where $I$ and $T$ are the paired image and description, and $i \in \{\mathbf{ICM}, \mathbf{FIR}\}$. $G^i$ represents one of the two generation stages, which consists of MTFG module and the generator corresponding to $i$.
With the help of TLU, our model can get a salient weight map for the manipulated features, forcing the model to reconstruct the image with manipulated features which come from the memories. Therefore, the latent memories can be learned with supervision in this stage. 

\textbf{Pseudo Ground-truth Feature Matching Loss.} To fully use the information of the paired data, we fuse the given image feature with the weight map as a pseudo ground-truth features of manipulating objects. With the pseudo ground-truth features, we can supervise the global texture feature $\bar{h}$ by applying a $L_1$ loss between them:

\begin{align}
    \mathcal{L}_{p} = \mathbb{E}_{I;T}[||
    \sum(v^i\odot\alpha)_{xy} / \sum\alpha_{xy} - \bar{h}||_1],
\end{align}
where $v^i$ is the feature map of the given image $I$,  $x \in [0, N-1]$ and $y \in [0, M-1]$ with $(N,M)$ being the size of the feature map $v^i$; $\alpha$ and $\bar{h}$ are the weight map and global texture feature predicted in the ICM stage with the matched image-description pair $I$ and $T$. 
%\jf{JF: (15) doesn't seem right. $i$ is used in both subscript and superscript!} 

\textbf{Randomized Memory Training Loss.}   
 
Since the textural features are synthesize by weighted combination of the memory banks, a description with rare words or grammar can make a poor weight prediction, thus leading to a unrealistic manipulated image. 
In our model, to overcome this problem, we propose a randomized memory training loss in the reconstruction stage. Specifically, we randomly select a set of attention weights $\Tilde{a}$ to fuse a new word-level texture feature in Eq.~(\ref{fuse}) and manipulate the image base on it. Then the manipulated image is fed into the reality discriminator. Since this memory does not involve the description, the discriminator should classify it as a real image to force each memory in the set can be decoded into a real image.

With all the memory vectors being in the real domain, the texture feature which is the convex combination of the memories are more likely to be in the real domain. In this way, the model can reduce the impact of noise and increase its robustness when an unusual but valid description is given. The memory training loss can be shown as, 
\begin{align}
    \mathcal{L}_{m}^i = -\mathbb{E}_{I;\Tilde{a}\sim P_\mathcal{a}}[log(D_I^i(G^i(I,\Tilde{a}))],
\end{align}
where $P_\mathcal{a}$ is the uniform distribution of random sampling operation in the memory set,
$I$ is the given image and $\Tilde{a}$ is the randomly sampled attention weights.

\subsubsection{\textbf{Adversarial Training}}

In this stage, adversarial losses are applied to force the manipulated images to be realistic and text-conform. During the adversarial training, the memories are kept frozen to avoid the influence of the noise from the random data sampling. Similar to \cite{goodfellow2014generative, xu2018attngan,nam2018tagan,li2020manigan}, the optimization consists of two steps and updates the model alternately.

In the first step, the discriminators of the two manipulation stages are updated at the same time. The losses are,
\begin{equation}
\begin{aligned}
    \mathcal{L}_{D}^i &= 
    -\mathbb{E}_{I,T,\hat{T}}[\log(D_I^i(I)) + \log(1 - D_I^i(G^i(I, \hat{T}))) \\
    &+ \log(D_T^i(I,T)) + \log(1 - D_T^i(G^i(I, \hat{T}), \hat{T}))]
\end{aligned}
\end{equation}
where $I$, $T$ and $\hat{T}$ are respectively the images, the matched descriptions and the descriptions randomly sampled from the dataset; $i \in  \{\mathbf{ICM}, \mathbf{FIR}\}$ is the identifier for the manipulation stages. The first two terms are calculated for the reality discriminator training and the last two terms are for the text-conformity discriminator training.

In the second step, we evaluate the manipulated image again to update the generators with the updated discriminators.
The formulation contain two parts which supervise the reality and conformity of manipulated images respectively,
\begin{align}
    \mathcal{L}_{I}^i &= -\mathbb{E}_{I,\hat{T}}[\log(D_I^i(G^i(I, \hat{T})))], \\
    \mathcal{L}_{T}^i &= -\mathbb{E}_{I,\hat{T}}[\log(D_T^i(G^i(I, \hat{T}), \hat{T}))],
\end{align}
where $\mathcal{L}_{I}^i$ is the reality loss and $\mathcal{L}_{T}^i$ is the text-conformity loss.

\subsubsection{\textbf{Integrated Loss}} The integrated loss for the generator is the weighted sum of all the generator related losses,
\begin{equation}
\begin{aligned}
\mathcal{L}_{G} &= \lambda_p\mathcal{L}_p \\
    &+ \sum_{i\in \{ \text{ICM, FIR}\}}
    \lambda_I^i\mathcal{L}_{I}^{i} +
    \lambda_T^i\mathcal{L}_{T}^{i} +
    \lambda_{rec}^i\mathcal{L}_{rec}^{i} +
    \lambda_m^i\mathcal{L}_{m}^{i},
\end{aligned}
\end{equation}
where $\lambda$s are the hyper-parameters to control the relative importance of the losses. Similarly, the integrated loss for the two discriminators are
\begin{align}
    \mathcal{L}_{D} &= \sum_{i\in \{ \text{ICM, FIR}\}}
    \beta^i\mathcal{L}_{D}^{i},
\end{align}
where $\beta^i$ are the hyper-parameters for the two manipulation stages.

\section{Experiments}
\subsection{Datasets and Settings}
We evaluate our MIM-Net on four different datasets with caption annotations including Celeba dataset, CUB dataset, Oxford-102 dataset, and Deep-Fashion dataset. All the images are resized to $256 \times 256$ during training.

\noindent\textbf{Celeba dataset} contains 202,599 face images in the wild with 192,050 images annotated by~\cite{nasir2019text2facegan}. For each annotated image, it has one paired caption. We randomly select 153,640 image and caption pairs for training and the rest for testing.

\noindent\textbf{CUB dataset} is a fine-grained bird classification dataset with caption annotations. There are 8,855 training images and 2,993 testing images in 200 bird categories. For each image, it has 10 different paired captions. We follow TAGAN~\cite{nam2018tagan} to split the training and testing images.

\noindent\textbf{Oxford-102 dataset} has a total of 8,189 flower images of 102 classes, and each image has 10 paired text descriptions. We follow TAGAN~\cite{nam2018tagan} to split the training and testing images.

\noindent\textbf{Deep-Fashion dataset} \cite{liu2016deepfashion} totally has 79K images with the sentence descriptions given by~\cite{zhu2017be}. We randomly select 63,183 images for training and the rest 15,796 images for testing.

\noindent\textbf{Implementation Details.} We use a batch size of 30 with Adam optimizer with $\beta$ = 0.999 and a fixed learning rate of 0.0002 to train the whole MIN-Net. We train 600 epochs for CUB and Oxford-102 dataset, 40 epoch for CelebA dataset, and 70 epochs for Deep-Fashion dataset. Our MIM-Net is able to output both $128 \times 128$, and $256 \times 256$ resolution images.

\begin{table}
\centering
\begin{tabular}{l|l|l|l|l}
\hline
Mthods  & IS & Sim &Diff &MP\\ \hline \hline
SISGAN  \cite{dong2017semantic}              & 2.24             & 0.045         &0.508          &0.022 \\ 
TAGAN \cite{nam2018tagan}                    & 3.32             & 0.048         &0.267          &0.035\\ 
ManiGAN \cite{li2020manigan}                 & \textbf{8.47}    & 0.101         &0.281          &0.072\\ \hline
Ours w/o $\mathcal{L}_{m}$,$\mathcal{L}_{p}$ & 4.38 $\pm$ 0.22  & 0.152             & 0.188             & 0.123\\ 
Ours w/o $\mathcal{L}_{m}$                   & 4.53 $\pm$ 0.17  & 0.154             & \textbf{0.187}             & 0.125\\ 
Ours 128                                     & 4.43 $\pm$ 0.29  &0.127          & 0.292         & 0.091\\ 
Ours 256                                     & 4.55 $\pm$ 0.31  &\textbf{0.171} &0.190 &\textbf{0.139}\\  \hline
Real Image                                   & 5.24 $\pm$ 0.30  &0.296          & 0             & 0.296\\ \hline
\end{tabular}
\caption{Quantitative results with state-of-the-art methods on CUB dataset. Compared with other methods, our entire model can achieve best performance in Sim and MP.  } 
\label{IS}
\end{table}

\begin{figure}%[t]
\centering
    \includegraphics[width=0.8\linewidth]{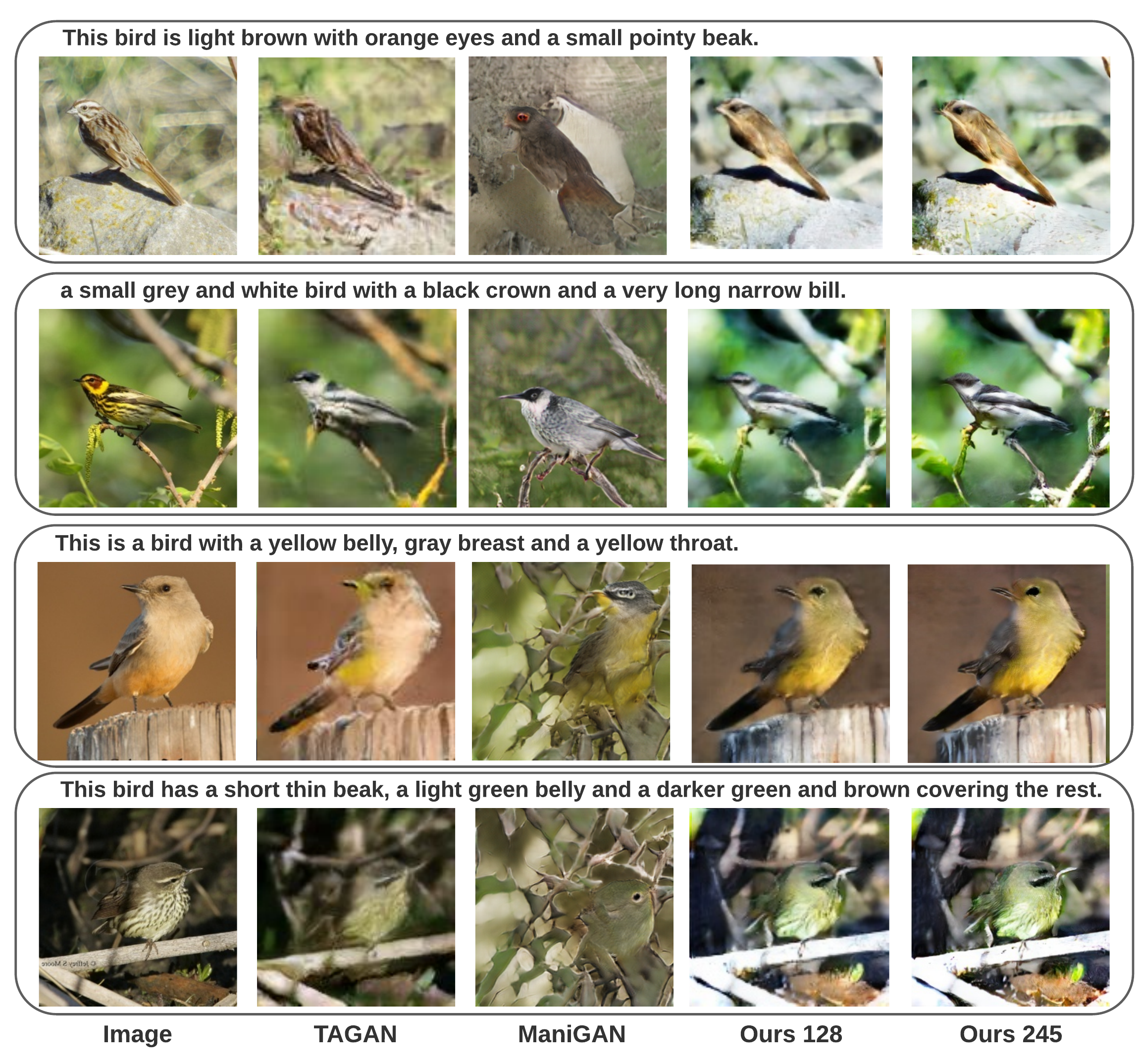}
    \caption{Visualization results for comparisons with existing methods. In contrast, our method can generate a text-conformed manipulation with a better structure and background preservation. }
    \label{compare}
\end{figure}

\begin{figure*}%[t]
\centering
    \includegraphics[width=0.8\linewidth]{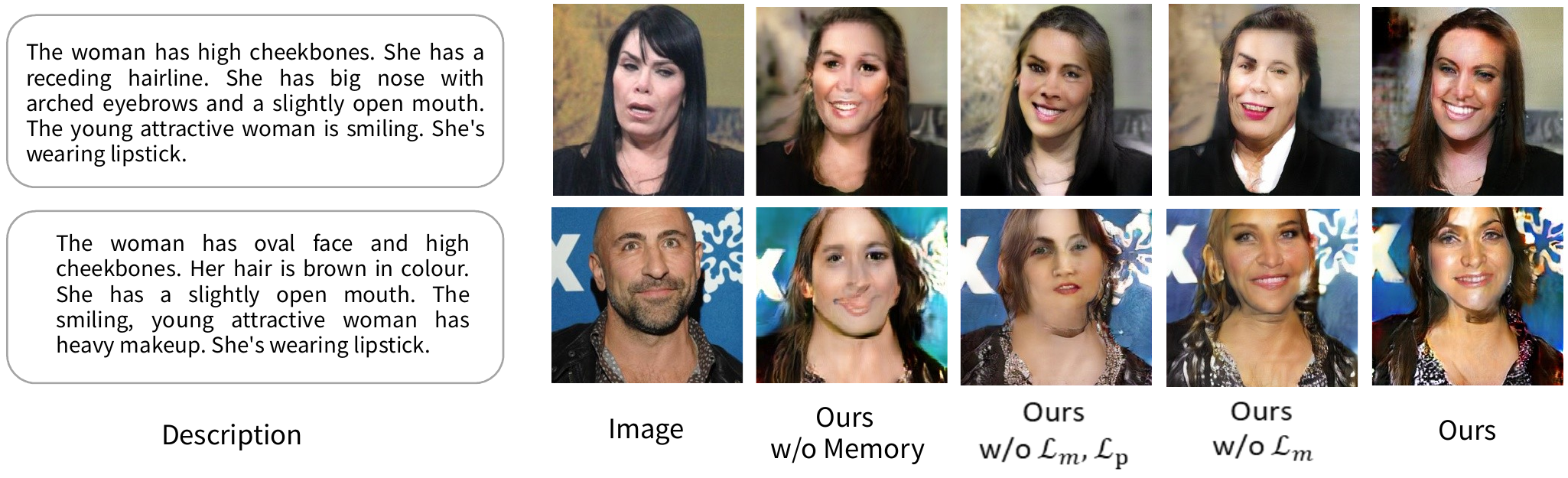}
    \caption{Visualization results for ablation study. We add our proposed memory and losses step by step with the results show in the 2nd-5th columns.} 
    \label{ablation}
\end{figure*}

\begin{figure}%[t]
\centering
    \includegraphics[width=\linewidth]{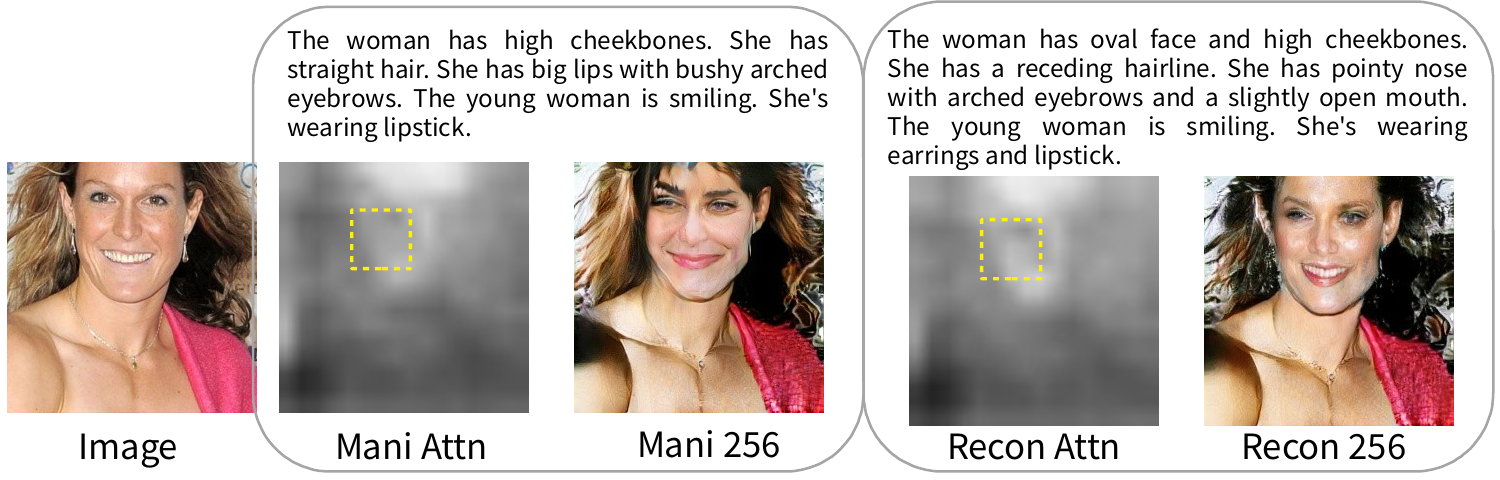}
    \caption{Visualization results for the Target Localization Unit (TLU). 2nd column: results of the manipulation (`Mani') stage with a description randomly selected from the dataset. 3rd column: results of the reconstruction (`Recon') stage with the paired description.}
    \label{recon}
\end{figure}

\begin{figure}
\centering
    \includegraphics[width=0.8\linewidth]{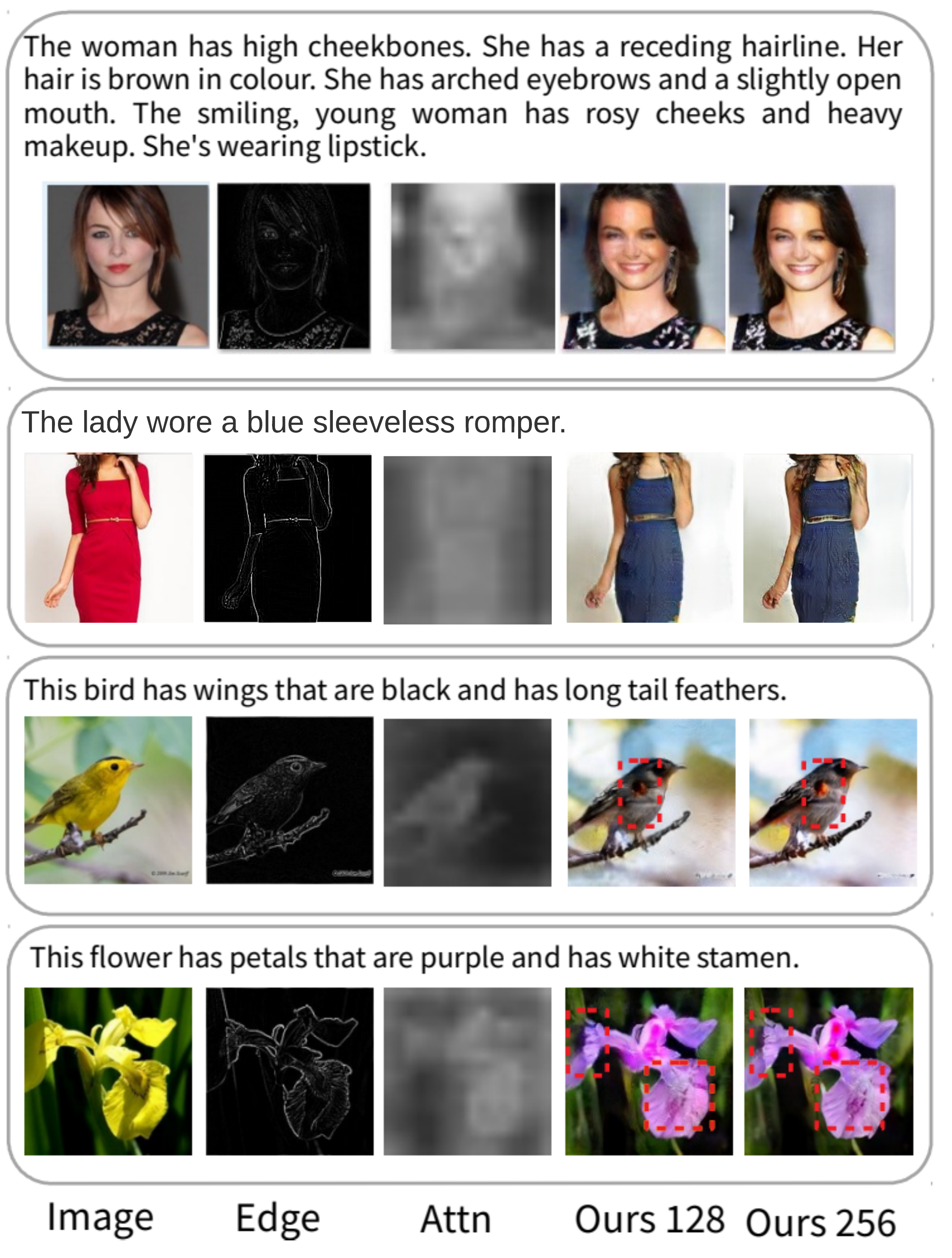}
    \caption{Visualization results for different test datasets. We can see that our model can localize manipulation regions and generate a coarse manipulated image and a refined one in the 4th and 5th columns, respectively. }
    \label{show_result}
\end{figure}

\begin{table*}
\centering
\begin{tabular}{l|l|l|l}
\hline
Mthods &\textbf{Semantic Relevance} &\textbf{Background Quality} &\textbf{Image Quality}                          \\ \hline \hline
ManiGAN \cite{li2020manigan}  & 3.44 $\pm$ 1.20 &3.01 $\pm$ 1.18 &3.28 $\pm$ 1.17                                   \\ \hline
Baseline (w/o Memory) & 2.29 $\pm$ 1.30 &2.24 $\pm$ 1.36 &2.30 $\pm$ 1.34                                           \\ \hline
Ours 256              & \textbf{3.50 $\pm$ 1.13} &\textbf{3.42 $\pm$ 1.21} &\textbf{3.51 $\pm$ 1.16}         \\ \hline
\end{tabular}
\caption{Human evaluation results on CUB dataset. } 
\label{HE}
\end{table*}

\subsection{Quantitative Results} 

Following ManiGAN~\cite{li2020manigan}, to demonstrate our proposed method is able to generate a realistic and semantic-matched manipulated image, we adopt the inceptions score (IS), Image-Text Similarity (Sim), Pixel Difference (Diff) and Manipulative Precision (MP) metrics as quantitative measures. Same as ManiGAN~\cite{li2020manigan}, we test our results on CUB dataset for fair comparison. Table~\ref{IS} shows the results, where the results of the comparing methods are directly taken from ManiGAN~\cite{li2020manigan}. It can be seen that our performance is improved by adding different training losses step by step, from IS 4.38 to 4.53 and to 4.55.
Compared with the state-of-the-art methods, our method has surpassed SISGAN and TAGAN significantly.
Although ManiGAN has a higher IS score, our IS is closest to the real images, which suggests that our generated images are more similar to the real images.

To calculate the Image-Text Similarity (Sim), following ManiGAN~\cite{li2020manigan}, we extract the manipulated image and supported text features using the pretrained encoders of AttnGAN~\cite{xu2018attngan}, and then compute the cosine similarity of image-text pairs to get the Sim result for each sample. Compared with the existing methods, our method improves Sim by 0.70, 0.123 and 0.126 than ManiGAN, TAGAN and SISGAN,  respectively, which indicate that our memory based manipulation can achieve better text-conformity for the manipulated images. 
  
Pixel Difference (Diff) is used to evaluate the pixel-level change between images before and after manipulation. The background which is not mentioned in the description is expected unchanged during manipulating, for which the pixels in the background should have lower Diff score. Since background usually takes up the majority region of images, lower Diff score thus suggests better background preservation. As shown in the third column of Table~\ref{IS}, our method reduces Diff significantly, from 0.281 to 0.190,  compared with the  state-of-the-art ManiGAN. This indicates that our method can better preserve the background.   
Manipulative Precision (MP) is proposed in ManiGAN~\cite{li2020manigan} to combine Sim and Diff to better reflect manipulation quality, which is defined as
\begin{equation}
MP = (1-Diff) \times Sim, 
\end{equation}
where higher $1-Diff \in [0,1]$ suggests better background preservation and higher Sim suggests better text-conformity. 
Thus, higher MP indicates better semantic-matching and background-preservation manipulation. Table~\ref{IS} shows that our method achieves a significant improvement in MP, increasing from 0.072 to 0.139,  compared with ManiGAN . 
\subsection{Qualitative Analysis}
To give a qualitative evaluation, we compare our visual results with the two SOTA methods, TAGAN and ManiGAN, with some examples given in Fig.~\ref{compare}. 
It can be seen that ManiGAN alters the background together with the object manipulation. However, shown in the fourth column, our results  can preserve better background and avoid generating the texture which is not provide by the supports. Compared with the results of TAGAN in the second column, we can see that our results are more realistic and clearer.

Moreover, our model can achieve a better structure preserving. As shown in the first sample, our result can keep a clearer boundary for both foreground and background and avoid them being distorted (like the rock in TAGAN and the bird in ManiGAN). In the second sample, ManiGAN edit the bird from a ``small'' one to a ``fat'' one. In contrast, our model can successfully keep the shape of the bird while matching the attributes in the description.
Our results are also more text-conformed. In the fourth example, the result of TAGAN is blurred and contains regions that match the textures of original bird, like crown and belly of the bird. In our result, the bird is generated mainly following the descriptions with the green belly and dark green at the end of feathers. 

We also conducted a human evaluation to evaluate the quality of the manipulated results on CUB dataset. To generate the test input pairs, we shuffle the test dataset and randomly select 30 images and captions from it to form the new test set for testers. For each set, we provide manipulated $256\times 256$ images from our final model, the baseline without memory and ManiGAN~\cite{li2020manigan} without annotations to testers for comparisons. The input image-description pairs are also provided as a reference for evaluation. The testers are required to evaluate the quality of the manipulated images in the following three aspects. 
\begin{itemize}
\item \textbf{Semantic Relevance} is to evaluate how the manipulated images are matched with the reference descriptions. 
 \item \textbf{Background Quality} is to measure the background preserving, i.e. how the background of manipulated images match with their supported images.
\item \textbf{Image Quality} is to evaluate how realistic the manipulated images are. 
 
\end{itemize}
For each aspect, we have 5 ranks from ``1'' to ``5'', representing the evaluation ratings from ``Very Poor'' to ``Very Good''. Thus, the higher the values, the better the performance in the evaluation. 

Totally 67 testers took this evaluation. The average ratings of the three aspects are summarized in Table~\ref{HE}. From the table, we can find that our method achieves better performance in all aspects,  compared with ManiGAN~\cite{li2020manigan} and our baseline. The gains are particularly significant in the aspects of background quality and image quality, which indicate our results are better background-preserving and more realistic.

\subsection{Ablation studies}

Table~\ref{IS} already shows the quantitative improvements of adding different losses. Fig.~\ref{ablation} further gives a qualitative ablation study. 
Comparing the models with and without memory, we can see that the model without memory cannot generate realistic local details like mouth (see 3rd column). Comparing ``Ours w/o $\mathcal{L}_{m}, \mathcal{L}_{p}$'' and ``Ours w/o $\mathcal{L}_{m}$'', we can find that the pseudo ground-truth loss $\mathcal{L}_{p}$ can better preserve the structure of the object. Comparing ``Ours w/o $\mathcal{L}_{m}$'' and ``Ours'', we can see that without $\mathcal{L}_m$, some blank regions are observed around the neck and cheek regions in the first and second examples, respectively. 

Fig. \ref{recon} illustrates the effect of our proposed TLU in both the reconstruction and manipulation stages. %It
Compared with the yellow bounding box of the weight map in the manipulation stage, the one in the reconstruction stage pays more attention to the cheek of the woman to learn the attribute of ``oval face'', which appears in the paired description but not in the manipulating one.
In this way, the model can ignore the image feature and use memories to reconstruct the attentive image regions, which leads to more effective memory learning.

Fig.~\ref{show_result} shows the intermediate results of MIM-Net during testing. 
The first and second columns are the original images and their boundary maps. The third column is the weight map predicted by TLU in the ICM stage. From the figure, we can observe that our TLU can generate salient weight maps for the manipulation regions during testing. For example, in the first example, our predicted weight map accurately highlights the regions matched with the description, such as ``mouth'', ``hairline'' and ``cheek-bones'', while ignoring ``eyes'', and ``nose'' that are not mentioned in the description. 
 
In the second sample, besides the region of the body, the map also covers arms and shoulders where the sleeves are in the original image to manipulate the dress from ``short-sleeved'' to ``sleeveless''. It indicates that the memory has the ability to learn some attributes beyond texture features like ``long-sleeved'' and ``sleeveless''. 

The last two columns in Fig.~\ref{show_result} are our manipulated results with a resolution of $128\times 128$ and $256\times 256$, respectively. It can be seen that  
the coarse images often contain less local details, like the fuzzy teeth and hair, while via FIR, the results are clearly improved, with clearer teeth, eyes, and hair.
In the fourth example, the FIR module restores the broken patches at the edges of the petal, making the manipulated image more realistic.

\section{Conclusion}
In conclusion, we have proposed a Memory-based Image Manipulation Network (MIM-Net). Different from the previous methods, we proposed to disentangle the image features into texture and structure parts and introduce a set of latent memories to represent the texture information. To learn accurate texture memories, we proposed a reconstruction stage with the novel weight maps to force the model to reconstruct the image with memory features. Moreover, we further introduced a memory-level adversarial training loss to keep the memories robust and prominent. With the proposed modules, our model is able to achieve better background preservation and robust text-conformed attribute manipulation on the four tested datasets.
\bibliographystyle{ACM-Reference-Format}
\bibliography{sample-base}
\end{document}